\documentclass[journal]{IEEEtran}
\usepackage[final]{graphicx}
\usepackage{amssymb}
\usepackage{url}
\usepackage{amsmath}
\usepackage{array}
\usepackage{algorithm}
\usepackage{algorithmic}
\usepackage{booktabs}
\usepackage{setspace}
\usepackage{multirow}
\usepackage{color,xcolor}
\usepackage{cite}
\usepackage{multirow,makecell}
\usepackage{diagbox}
\usepackage{balance}
\usepackage{amssymb}
\usepackage[nameinlink,capitalise]{cleveref}
\usepackage{caption}
\usepackage{overpic}

\DeclareGraphicsExtensions{.pdf,.jpg,.png}
\graphicspath{{./figures/}}
\graphicspath{{figures/}}

\definecolor{red}{rgb}{0.8,0,0}

\newcommand{\figref}[1]{Fig.~\ref{#1}}
\newcommand{\tabref}[1]{Tab.~\ref{#1}}
\newcommand{\secref}[1]{Sec.~\ref{#1}}

\newcommand{\equref}[1]{Eq.~(\ref{#1})}

\def\ie{\emph{i.e.~}}

\def\etal{{\em et al.~}}

\begin{document}

\title{A Compact Neural Network-based Algorithm for Robust Image Watermarking}

\author{Hong-Bo Xu$^{*}$, 
        Rong Wang$^{*}$,
        Jia Wei,
        and Shao-Ping Lu, ~\IEEEmembership{Member,~IEEE} 
\thanks{Hong-Bo Xu, Rong Wang, Jia Wei and S.-P. Lu are with TKLNDST, CS, Nankai University, China. Shao-Ping Lu is the corresponding author (e-mail: slu@nankai.edu.cn). $^{*}$ denotes equal contribution. } 
}

\maketitle
\begin{abstract}


Digital image watermarking seeks to protect the digital media information from unauthorized access, where the message is embedded into the digital image and extracted from it, even some noises or distortions are applied under various data processing including lossy image compression and interactive content editing.
Traditional image watermarking solutions easily suffer from robustness when specified with some prior constraints, while recent deep learning-based watermarking methods could not tackle the information loss problem well under various separate pipelines of feature encoder and decoder.
%
In this paper, we propose a novel digital image watermarking solution with a compact neural network, named Invertible Watermarking Network (IWN).
%
%
Our IWN architecture is based on a single Invertible Neural Network (INN), this bijective propagation framework enables us to effectively solve the challenge of message embedding and extraction simultaneously, by taking them as a pair of inverse problems for each other and learning a stable invertible mapping.
In order to enhance the robustness of our watermarking solution, we specifically introduce a simple but effective bit message normalization module to condense the bit message to be embedded, and a noise layer is designed to simulate various practical attacks under our IWN framework.
%
%
Extensive experiments demonstrate the superiority of our solution under various distortions.
    
\end{abstract}


%
\IEEEpeerreviewmaketitle


\section{Introduction}


\IEEEPARstart{W}{ith} the rapid development of digital information processing technologies, various digital media content have been widely used in many areas. 
Because digital media is easy to propagate, copy and modify, how to protect the copyright of digital media has become a crucial but also practical problem. 
Digital watermarking aims to solve this problem by embedding extra information into the digital media and extracting such extra data for authorized access.
Nowadays, digital watermarking has been widely used in many applications include broadcast monitoring~\cite{chen2001quantization}, copy control~\cite{faundez2007speaker}, and device control~\cite{broughton1989interactive}.
In this paper, we focus on digital image watermarking.
In particular, the expected image watermarking algorithm asks for embedding the message (\ie watermark) into the cover image (\ie the image requiring authorized access) to obtain the watermarked image. 
In addition, the image watermarking algorithm needs to recover the original message as much as possible from the watermarked image.
%

Although digital image watermarking has been widely studied in the academic community, it remains a challenging issue.
There are three key factors to measure the performance of the digital image watermarking algorithm, namely the robustness, imperceptibility, and capacity.
The robustness requires the message embedded into the image to survive under malicious and non-malicious attacks. 
The imperceptibility needs the watermarked image to be as identical as possible to the original one, and it emphasizes that when changing the original image it is negligible for people to detect such activity.
The capacity refers to the amount of messages that can be embedded.
Besides that, the security~\cite{loan2018secure} and complexity~\cite{Pexaras2019OptimizationAH} aspects are also considered under certain conditions in the expected watermarking scheme, although in many cases they are with much less priority.
Moreover, these key factors are conflicted between each other, and it is impossible to satisfy these features simultaneously~\cite{tao2014robust}. 
Existing applications of watermarking algorithms usually focus on some special features or intend to make a trade-off for the above-mentioned features.
For instance, watermarking for copyright protection considers the better robustness, while watermarking for broadcast monitoring concerns a larger capacity.
For most existing deep learning-based robust image watermarking systems~\cite{zhu2018hidden,tancik2020stegastamp,luo2020distortion}, the better robustness and imperceptibility are more important, but how to make a trade-off to satisfy those conflict features of watermarking is still one of the main challenges in this research domain.

\begin{figure}[tb]
    \centering
    \includegraphics[width=1.0\linewidth]{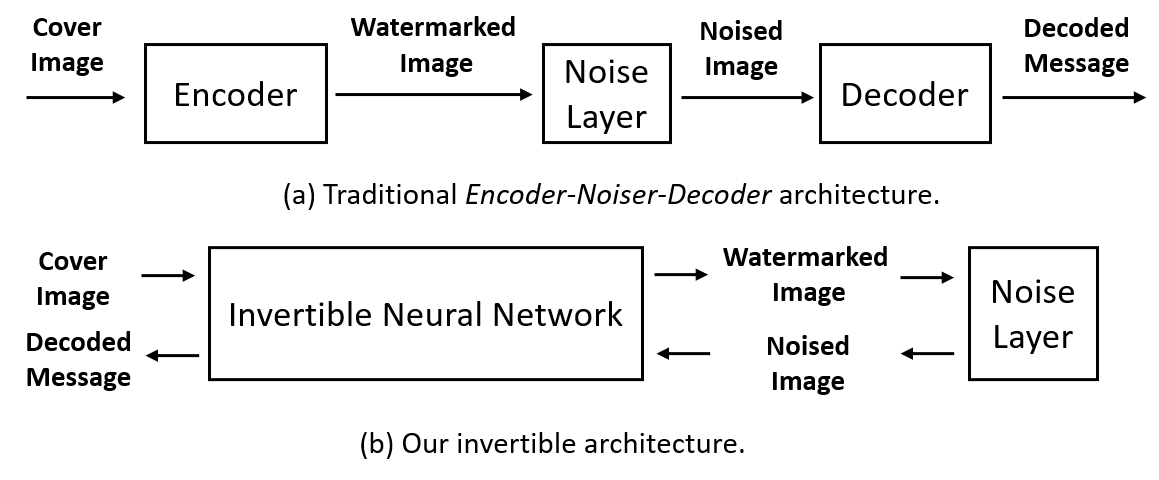}
    \caption{Illustration of traditional Encoder-Decoder solutions and our compact architecture.}
    \label{our_work}
\end{figure}

Traditional image watermarking techniques usually embed watermarks in the spatial domain or frequency domain~\cite{ray2020recent}.
For the spatial domain-based watermarking techniques, one of the advantages is the computational efficiency when directly changing pixel values of the image, while it easily suffers from robustness.
On the contrary, frequency domain-based watermarking solutions would obtain higher robustness by manipulating frequency coefficients of the image in the frequency domain, and they are usually with higher computational complexity.
%
The main drawback of traditional watermarking methods is that they are specified on some prior constraints or targets, making them difficult to be generalized for novel types of attacks~\cite{byrnes2021data}. This significantly constraints them in some limited applications.
In recent years, deep neural networks have already been applied to digital image watermarking~\cite{KANDI2017247, zhu2018hidden, Wen2019ROMarkAR, wengrowski2019light, tancik2020stegastamp, jia2020rihoop, yu2020attention, zhong2020automated}.
Because of the strong representation abilities of deep neural networks, these approaches have achieved better robustness and imperceptibility than traditional methods.
In addition, neural networks can be retrained to resist novel types of attacks, or to focus on particular features such as the robustness and imperceptibility without designing a new specialized sophisticated algorithm, enabling them possible to develop an adaptable and generalized framework for various watermarking applications~\cite{zhu2018hidden}.
However, most of them use the Encoder-Noiser-Decoder framework~\cite{zhu2018hidden, wengrowski2019light, tancik2020stegastamp, jia2020rihoop, yu2020attention, zhong2020automated}, as shown in \figref{our_work} (a).
In general, this framework employs a separate encoder and decoder to embed and extract watermark respectively.
This asks for careful construction for both message embedding and extraction, and the training of two separate neural networks needs complicated parameter tuning.

In this paper, we propose a novel digital image watermarking scheme named invertible watermarking network (IWN) using invertible neural network (INN).
Inspired by that from the perspective of reversible image conversion (RIC), INN can alleviate the information loss problem better than classic neural network architecture~\cite{cheng2021iicnet}, we thus consider watermark embedding and extracting as a pair of inverse problems, and we effectively solve them with INN.
Different from existing Encoder-Decoder based deep watermarking networks, our compact IWN respectively applies watermark embedding and extracting in the forward and reverse process of INN sharing all network parameters, as shown in \figref{our_work} (b).
As already demonstrated that INN is an effective tool for embedding and extracting a large amount of information~\cite{lu2021large}, our IWN achieves high imperceptibility benefited from the strictly invertible property of INN~\cite{INN_2018analyzing}.
To enhance the robustness, we introduce a well-designed bit message normalization module and a noise layer in our system.
The former also ensures that different lengths of the bit messages can be easily adapted with a high recovery accuracy in our IWN.
With the noise layer which is used to simulate various attacks, the strong fitting ability of our IWN enables us to effectively learn the robustness against various practical distortions.
%
%
Extensive experiments show that our method achieves better results than the most commonly used baseline.
In addition, we are the first to introduce INN into the field of watermarking, and we hope to enlighten the follow-up research.

In summary, the main contributions of this paper are:
\begin{itemize}
    \item To our knowledge, we are the first to introduce invertible neural networks into digital watermarking, and we propose an invertible watermarking network (IWN) for robust and blind digital image watermarking.
    \item We introduce a bit message normalization module for condensing the messages and a noise layer for simulating various attacks, respectively, with which the watermarking robustness is significantly improved. 
    \item We provide extensive experiments to demonstrate the superiority of our method under a variety of distortions.
\end{itemize}

The rest of this paper is organized as follows. We review the related work in \secref{sec_related_work}. The proposed method is described in \secref{sec_proposed_method}, followed by extensive experiments in \secref{sec_experiments}. Finally, the conclusion and future work are given in \secref{sec_conclusion}.
\section{Related Work} \label{sec_related_work}
Since the terminology digital watermarking first appeared in~\cite{van1994digital}, it has been an active research area~\cite{katzenbeisser2000digital, cox2002digital, cox2007digital} with many applications such as copyright protection and owner identification. 
Besides natural images, digital watermarking has also been used in other fields like medical image watermarking~\cite{haddad2020joint}, video watermarking~\cite{asikuzzaman2014imperceptible}, dynamic software watermarking~\cite{ma2019xmark}, 3D watermarking~\cite{gao2021thermotag, hou2017blind}, audio watermarking~\cite{liu2018patchwork}, neural network watermarking~\cite{wu2020watermarking} and so on. 
In this paper, we focus on robust digital image watermarking and we briefly review two main research areas that are most relevant to our work, \ie digital image watermarking and invertible neural networks, in this section.

\subsection{Digital Image Watermarking}

Traditional digital image watermarking techniques usually embed messages in spatial domains or frequency domains~\cite{ray2020recent}. 
In general, those methods of the spatial domain directly embed watermarks by manipulating bitstreams or pixel values~\cite{su2018robust, li2013general, deng2010local}.
Among them, Least-Significant-Bit (LSB)~\cite{van1994digital} is a representative work of this subcategory.
However, it easily suffers from low capacity and sensitivity to various image processing attacks~\cite{byrnes2021data}.
On the other hand, frequency domain-based watermarking techniques modify the frequency coefficients when embedding watermarks.
Compared with the spatial domain-based watermarking methods, these solutions further improve the robustness, imperceptibility, capacity, fidelity, and security with the cost of higher computational complexity~\cite{2013dabas, 2018kumar}.
In this class of watermarking methods, the commonly used frequency domains include Discrete Cosine Transform (DCT) domain~\cite{huang2019enhancing}, Discrete Fourier Transform (DFT) domain~\cite{hamidi2018hybrid}, Discrete Wavelet Transform (DWT) domain~\cite{guo2002digital} and contourlet domain~\cite{bao2005image, bi2007robust}. 
For instance, Kang~\etal\cite{kang2003dwt} propose to embed the spread-spectrum watermark in the coefficients of the LL subband in the DWT domain, and Sadreazami~\etal\cite{sadreazami2018robust} design to embed the watermark in the contourlet domain.
They observe the robustness against JPEG compression of the low frequency component in the wavelet domain and the contour component in the contourlet domain, respectively.
This excellent idea of finding robust invariant under attacks is also utilized to resist geometric distortions including translation, rotation, and cropping~\cite{zhang2011affine, pereira2000robust, xiang2008invariant, tian2013ldft}.
The main drawback of these traditional watermarking methods is that they are specified on some prior constraints or targets, making them difficult to be generalized for novel types of attacks~\cite{byrnes2021data}.
In other words, these techniques can only handle some limited tasks.

Recently many researchers apply neural networks to digital image watermarking, and indeed some novel methods bring superior robustness and imperceptibility over traditional methods.
For example, Kandi~\etal\cite{KANDI2017247} first introduce convolutional neural networks (CNNs) to non-blind watermarking.
Mun~\etal\cite{mun2017robust} further propose a blind watermarking architecture based on CNN to embed and extract watermarks.
Zhu~\etal\cite{zhu2018hidden} propose an end-to-end neural network with adversarial training for both steganography and robust blind watermarking. 
ROMark~\cite{Wen2019ROMarkAR} simplifies adversarial training by using a min-max formulation for robust optimization. 
After that, RedMark~\cite{ahmadi2020redmark} uses two Fully Convolutional Neural Networks (FCNs) with residual connections to embed watermarks in the frequency domain without adversarial training. 
Different from the dependent deep hiding methods (DDH)~\cite{ wengrowski2019light, tancik2020stegastamp, jia2020rihoop, yu2020attention, zhong2020automated}, which adapt the watermark to the original cover image, UDH~\cite{zhang2020udh} proposes a universal deep hiding method to embed the watermark independent of the cover image. 
These existing works have demonstrated a variety of neural network structures to effectively realize the message embedding and extraction, ensuring that the watermarked image and the cover image have little or even no perceptual differences.

For existing deep learning-based watermarking methods, the noise layer is usually introduced to the networks for dealing with various distortions.
However, in order to train the entire network in an end-to-end manner, the noise layer must be differentiable. 
For non-differentiable distortions including JPEG compression, some methods~\cite{zhu2018hidden,ahmadi2020redmark,tancik2020stegastamp} turn to simulate them with a differentiable approximation, allowing the network to be trained in an end-to-end style.
In~\cite{luo2020distortion} and~\cite{wengrowski2019light}, some distortions are generated by a trained CNN instead of explicitly modeling distortions from a fixed pool during training, which is another way to deal with non-differentiable and hard modeled distortions.
In addition, Liu~\etal\cite{liu2019novel} design a redundant two-stage separable deep learning framework to address the problems in one-stage end-to-end training, such as image quality degradation and difficulty to simulate noise attacks using differentiable layers. 
Although many strategies have been proposed to deal with various distortions, how to ensure the robustness of the digital watermark in various situations is still a problem that needs to be solved well.

Besides the noise layer, most existing robust image watermarking systems based on deep learning use Encoder-Noiser-Decoder frameworks~\cite{zhu2018hidden, wengrowski2019light, tancik2020stegastamp, jia2020rihoop, yu2020attention, zhong2020automated}.
Among them, the encoder embeds the watermark into the cover image in an imperceptible manner,
and the decoder recovers the watermark message from the distorted watermarked image.
This kind of architecture usually asks for sophisticated designing of both the encoder and decoder, resulting in much complex training with carefully tuning parameters. 
Different from those previous works where the encoder and decoder are two independent networks, we adopt a bijective INN for watermark embedding and extraction. 

%

\subsection{Invertible Neural Network (INN)}
In recent years, INN has attracted much attention because of their efficient inversion.
INN is usually proposed for the flow-based generative model, where a stable invertible mapping is learned between the complex data distribution ${p}_X$ and a simple latent distribution $p_Z$. 
NICE~\cite{est_2014nice} and RealNVP~\cite{est_2016nvp} propose the additive and the affine coupling layers, respectively. These coupling layers are the basic component of INN, which satisfy the requirements of efficient inversion and a tractable Jacobian determinant. 
In~\cite{INN_2017understand,INN_2018analyzing} the explanation is specially explored for the invertibility. 
In~\cite{INN_mintnet}, flexible INN is constructed with masked convolutions under some composition rules.
An unbiased flow-based generative model is also introduced in~\cite{INN_residualflows}.
Besides, Glow~\cite{est_2018glow}, FJORD~\cite{est_2018ffjord}, i-RevNet~\cite{inn_2018revnet} and i-ResNet~\cite{INN_2019IRB} achieve better generation results by continuously improving the network representation capacity. 
\begin{figure*}[t]
    \centering
    \includegraphics[width=1.0\linewidth]{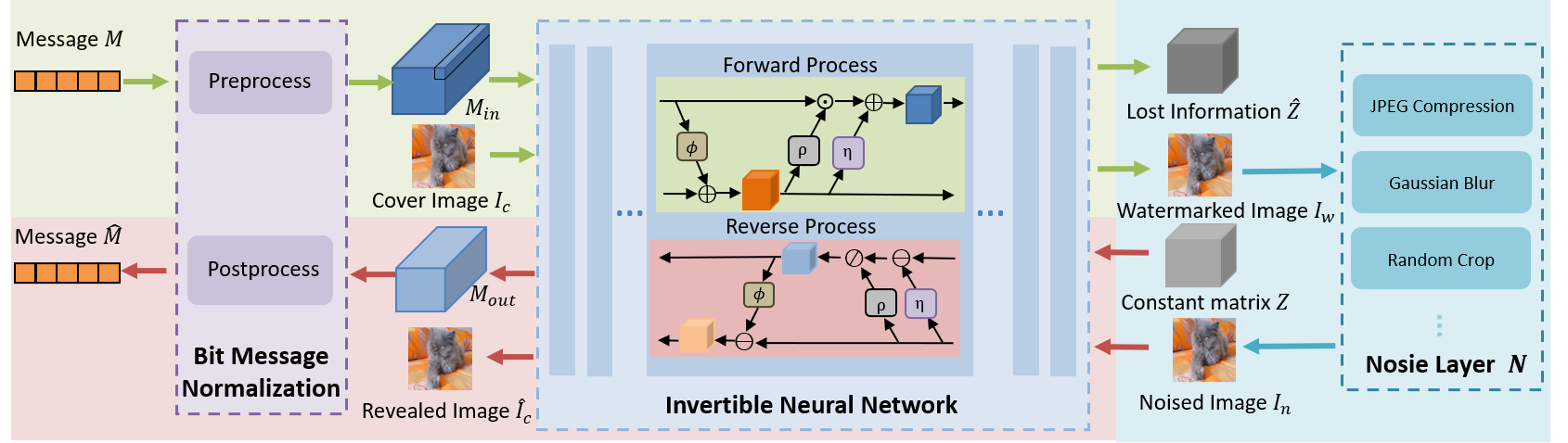}    \caption{Overview of our IWN, which contains a bit message normalization module, an invertible neural network (INN) and a noise layer. The preprocessed bit message $M_{in}$ and the cover image $I_c$ are fed into the bijective INN for embedding to obtain the watermarked image $I_w$. By introducing various noises and distortions, the noised image $I_n$ is served as the input of the INN's reverse mapping, the revealed message tensor $M_{out}$ and the image $\hat{I_c}$ are restored simultaneously. The watermark message $\hat{M}$ is finally obtained through postprocessing. }
    \label{over_view_fig}
\end{figure*}

In this context, INN has been used for a variety of challenging tasks due to their powerful fitting ability. 
For example, a conditional invertible neural network (cINN) is introduced for guided image generation~\cite{INN_guided}, including MNIST digits generation and image colorization. cINN is also used for network-to-network translation~\cite{rombach2020net2net} and image-to-video synthesis~\cite{Dorkenwald_2021_CVPR}. 
In addition, there are different solutions specified for image scaling~\cite{sr_2020eccv}, image compression~\cite{xie2021enhanced}, image or video super-resolution~\cite{app_2019residual}, image denoising~\cite{liu2021invDN}, underexposed image enhancement~\cite{zhao2021deep} and image color adjustment~\cite{zhao2021decolor}.
As the latest work, Cheng~\etal\cite{cheng2021iicnet} also propose a generic framework for the reversible image conversion, namely IICNet, which aims to encode a series of input images into a single image and decode them. %
Particularly, Lu~\etal\cite{lu2021large} firstly introduce INN into large-capacity image steganography, where up to 5 images are successfully embedded into a host image with the same spatial and color resolution. %
These latest advances demonstrate that INN has great potential in data embedding and extraction.
However, they all ignore the robustness issue that the embedded image may be manipulated under image compression and other distortion conditions.
On the contrary, our approach focuses on solving this robustness challenge.

\section{Proposed Method} \label{sec_proposed_method}

\subsection{Overview}

Instead of employing a cascading Encoder-Noise-Decoder architecture that is widely used in existing methods, here we propose an invertible watermarking network IWN, where a bijective INN is used to embed and extract the message. 
%
As shown in \figref{over_view_fig}, our compact IWN contains three components: 1) the invertible neural network, 2) bit message normalization module which includes the preprocessing and postprocessing sub-modules, and 3) noise layer $N$.
To efficiently represent the bit message, the bit message normalization module is used to convert the original bit sequence $M$ into a normalized tensor $M_{in}$. 
After that, INN is used as our backbone for efficient message embedding and extraction.
In this component, the forward process of INN takes the cover image $I_c$ and the preprocessed message $M_{in}$ as input, and it generates the watermarked image $I_w$ which is as similar as possible to the original image $I_c$ and $\hat{Z}$ which is lost information. 
The noise layer $N$ is then introduced to deal with various noises and distortions produced by practical image operations.
Our solution combines different noises to the watermarked image $I_w$ and obtains the simulated noised image $I_n$.
To extract the watermark message, the noised image $I_n$ is fed into the reverse process of INN to generate the output message tensor $M_{out}$ and the revealed cover image $\hat{I_c}$. 
With the message postprocessing sub-module of bit message normalization, we finally extract the bit message sequence $\hat{M}$ from $M_{out}$.
More details about the introduced notations are summarized in \tabref{notion}.

\begin{table}[tbp]
\centering
\footnotesize
    \caption{ Introduced notations. 
    }
    \label{notion}
\begin{tabular}{c|l }
\hline
\hline
 Notation & Description \\
\hline
    $I_c$ & Cover image \\ \hline
    $I_w$ & Watermarked image \\ \hline
    $I_n$ & Noised image produced by distortion simulation \\ \hline
    $\hat{I}_c$ & Revealed cover image \\ \hline 
    $M$ & Original bit message \\ \hline 
    $\hat{M}$ & Extracted bit message \\ \hline 
    $M_{in}$ & Preprocessed message tensor for INN's forward input \\ \hline 
    $M_{out}$  &  Revealed message tensor of INN's reverse output\\ \hline 
    $\hat{Z}$ & Lost information   \\ \hline 
    $Z$ & Constant matrix for recovering message  \\ \hline 
    
\end{tabular}
\end{table}
\subsection{Invertible Neural Network (INN)}
INN is powerful and effective in dealing with reversible problems, especially for data hiding and recovery~\cite{lu2021large}.  
Intuitively, watermarking is a special application of image steganography when a bit sequence message is taken as the hidden data. 
In this sense, the original bit message $M$ should be converted into a tensor $M_{in}$ with the same spatial resolution as the cover image by preprocessing sub-module of our bit message normalization.
%
%
After that, we use the forward process of INN for message embedding and its reverse process for message extraction. 
As shown in~\figref{over_view_fig}, in the forward process, 
the message tensor $M_{in}$ and the cover image $I_c$ are served as inputs, 
and the corresponding outputs are the watermarked image $I_w$ and a matrix $\hat{Z}$ which is just to satisfy the structural consistency of INN and will not be used for reverse mapping. 
As for the reverse process of our INN, the noised image $I_n$ and a predefined constant matrix $Z$ are fed in, then the revealed cover image $\hat{I}_c$ and the message tensor $M_{out}$ are extracted. Finally, the message $\hat{M}$ is obtained by the postprocessing sub-module of our bit message normalization. 
%

As shown in \figref{over_view_fig}, our INN consists of several invertible blocks with the same structure, and each block includes three sub-modules.
INN contains $b_1$ and $b_2$ two branches, corresponding to the hidden message and the cover image, respectively.  
For the $l$-th invertible block, the input is  [$b_1^l$ , $b_2^l$] and its output is [$b_1^{l+1}, b_2^{l+1}$], where $[\cdot,\cdot]$ is the concatenation operator in channel dimensions.
Formally, the forward process is calculated as follows:
\begin{equation}
  \label{eq:inn_ours}
  \begin{split}
    & b_2^{l+1} = b_2^{l} + \phi(b_1^l), \\
    & b_1^{l+1} = b_1^l \odot \mbox{exp}( \rho (b_2^{l+1}))  + \eta(b_2^{l+1}),
  \end{split}
\end{equation}
where $\phi(\cdot)$, $\rho(\cdot)$ and $\eta(\cdot)$ are convolution operations, $\mbox{exp}(\cdot)$ is the Exponential function and $\odot$ is the Hadamard product.
Accordingly, the reverse process in the $l$-th invertible block is calculated as follows:
\begin{equation}
  \label{eq:inn_ours_back}
  \begin{split}
    & b_1^l =( b_1^{l+1} - \eta(b_2^{l+1}) )\odot \mbox{exp}(-\rho (b_2^{l+1})  ) , \\
    & b_2^{l} =  b_2^{l+1} - \phi(b_1^l).
  \end{split}
\end{equation}
In other words, given  $b^{l+1}$ = [$b_1^{l+1},  b_2^{l+1}$], we can accurately calculate  $b^l$ =  [$b_1^{l},  b_2^{l}$] according to~\equref{eq:inn_ours_back}. 
By cascading, given the reverse input [$Z,I_n$], our output [$M_{out}, \hat{I}_c$] can be solved. 
It is worth noticing that the three sub-modules $\phi(\cdot)$, $\rho(\cdot)$ and $\eta(\cdot)$, which contain the learnable parameters, appear both in the ~\equref{eq:inn_ours} of forward process and the ~\equref{eq:inn_ours_back} of reverse processe. 
That is to say, INN shares all parameters during its forward and reverse mapping operations.
Benefiting from this architecture, INN performs stable and efficient inversion operations, which is exactly what we need in the watermarking task.

To optimize the network, we calculate the loss function for the above four items of outputs, respectively.
One of our goals is that there is no visual difference between the watermarked image $I_w$ and the original cover image $I_c$, so we introduce the loss function $\mathcal{L}_w$ to achieve that:
\begin{equation}
\mathcal{L}_w =  ||I_w - I_c||, 
\end{equation}
where $||\cdot||$ is the combination of $l_1$ norm and $l_2$ norm.
Similarly, we introduce $\mathcal{L}_{m}$ to ensure that $M_{out}$ and $M_{in}$ are as close as possible:
\begin{equation}
\mathcal{L}_m =  ||M_{out} - M_{in}||. 
\end{equation}
In addition, we also add the constraints $\mathcal{L}_{z}$ and $\mathcal{L}_{c}$ for the matrix $\hat{Z}$ and the revealed cover image $\hat{I}_{c}$, respectively:
\begin{equation}
  \label{eq:split_loss}
  \begin{split}
    & \mathcal{L}_z =  ||\hat{Z} - Z||, \\
    & \mathcal{L}_{c} = ||\hat{I}_{c}- I_c)||.
  \end{split}
\end{equation}
Finally, the total loss $\mathcal{L}_{total}$ of our system is formulated as:
\begin{equation}
  \label{eq:total_loss}
  \begin{split}
    \mathcal{L}_{total} = \omega_w \mathcal{L}_w + \omega_{m} \mathcal L_{m} + \omega_z \mathcal{L}_{z} + \omega_{c} \mathcal{L}_{c},
  \end{split}
\end{equation}
where $\omega_w$, $\omega_m$, $\omega_{z}$, $\omega_{c}$ are the weights of the corresponding losses presented above.
Note that under the Crop and Cropout attacks, we only calculate $\mathcal{L}_m$ in the cropped region, and multiply the corresponding ratio of the origin shape to the cropped region shape.
Please refer to \secref{sec_noise} for more details on how we deal with the Crop and Cropout attacks.
%

\subsection{Bit Message Normalization}
%
In general, there is a conflict between the robustness and the capacity of watermarking, \ie{,} when embedding more information, the watermarking scheme is more vulnerable to attacks.
Therefore, we introduce a novel bit message normalization module  which contains preprocessing and postprocessing sub-modules to normalize the bit message in a simple but effective way, with which the robustness is significantly improved.
Specifically, different with HiDDeN~\cite{zhu2018hidden} that uses one channel to represent each bit of message, our bit message normalization module can represent many bits with just one channel.
Moreover, this module allows us to flexibly adjust the watermarking capacity according to different practical applications, without changing the network architecture of our system. Here we introduce the details of our bit message normalization module.
\begin{figure}[t]
    \centering
    \includegraphics[width=0.95\linewidth]{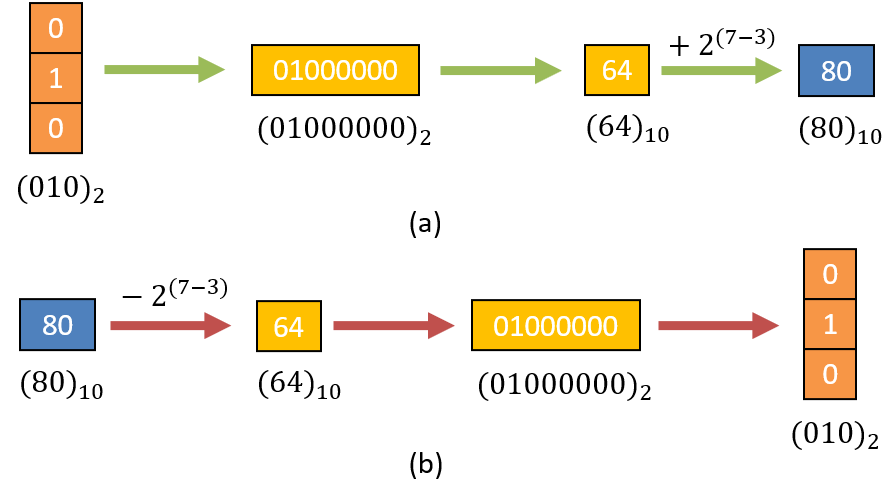}
    \caption{An example of our bit message normalization, where (a) and (b) are treated by the preprocessing and postprocessing sub-modules, respectively. Here the binary number $(010)_2$ is composed of 3 bits \{0, 1, 0\} from a message when $l/c=3$. Through left shifting $8-l/c=5$ bits and adding offset $2^{7-l/c}=16$, we obtain the decimal number $(80)_{10}$ to represent \{0, 1, 0\} in one channel. Through our postprocessing sub-module, the number of $80$ can be reversibly converted to 3 bits \{0, 1, 0\} accordingly.}
    \label{bit_process}
\end{figure}

\textbf{Preprocessing.}
The main purposes of this sub-module are to improve robustness and convert a sequence of bits $M$ into a tensor $M_{in}$ for the use of convolutional networks, including the following two steps: bit transformation and broadcasting.  
For a bit message sequence $M$ of length $l$, we divide it into $c$ groups. 
In each group, there exists $l/c$ $(\leq8) $ bits, which is treated as a binary number.
For the convenience of training, we convert the $c$ grouped binary numbers into the corresponding 8-bit integers aligned with the most common 8-bit color depth.
In other words, we transform the $c$ grouped binary numbers into their corresponding decimal numbers.
In order to encode the bit information into the highest bits for higher error tolerance rather than on lower bits positions, we specially left shift each binary number by $8-l/c$ bits.
Then by treating the shifted binary numbers as 8-bit integers, we add an offset $2^{7-l/c}$ on them, ensuring that the mean value of all $c$ integers generated from a random bit sequence equals to 128, which is the median of color pixel values between 0 and 255.
In~\figref{bit_process} (a) we show an example for encoding 3 bits into one channel, \ie{} $l/c=3$.
In order to spread the watermark message over all image pixels, the message with $1 \times 1\times c$ shape is then broadcast to the input message as $M_{in}$ ($H \times W\times c$), where $H$ and $W$ are the height and width of the cover image $I_c$, respectively.
Interestingly, the bit message processing method in HiDDeN~\cite{zhu2018hidden} can be regarded as a special case of ours when $l/c=1$. 
%
%
%

\begin{figure}[t]
    \centering
    \includegraphics[width=1.0\linewidth]{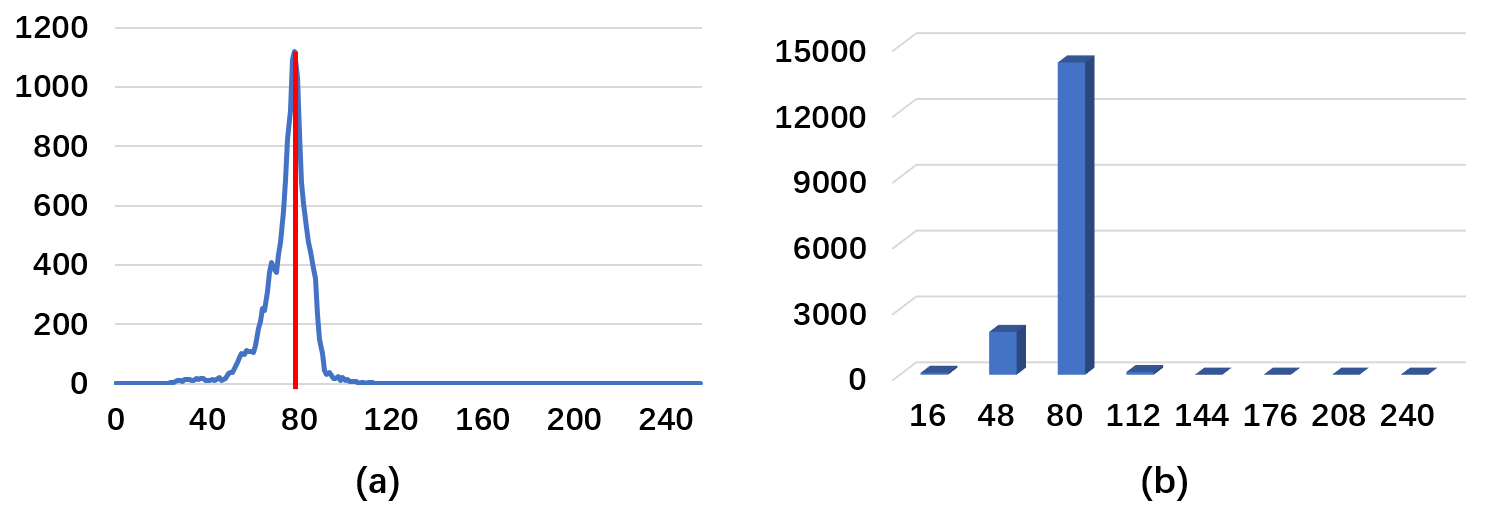}
    \caption{Data distribution statistics for a single channel of the revealed message tensor $M_{out}$ when $l/c=3$. (a) is the original data distribution in [0, 255], where the red line represents the ground-truth. (b) is the recovered result after quantizing them into the $2^{l/c} = 8$ categories. }
    \label{mode_fig}
\end{figure}

\textbf{Postprocessing.}
Our goal is to obtain a bit message sequence $\hat{M}$ from the tensor $M_{out}$ ($H \times W\times c$), so we need to get $c$ integers at first.
Obviously, for the $H\times W$ elements in each channel, we just need to map them to a single decimal number.
As shown in \figref{mode_fig} (a), the original data distribution of one channel from $M_{out}$ presents a single peak state.
To eliminate the interference of outliers for each channel, we convert all numbers to their corresponding nearest ground-truth values, which share $2^{l/c}$ candidates as shown in \figref{mode_fig} (b), and then we take the mode of converted numbers as the final extracted number.
After that, the $c$ integers are converted to a bit message sequence according to the inverse process of preprocessing, as shown in~\figref{bit_process} (b).  
Specifically, it includes offset subtraction, right shift, and binary conversion operations.

\subsection{Noise Layer} \label{sec_noise}
\begin{figure}[tb]
    \centering
    \includegraphics[width=1.0\linewidth]{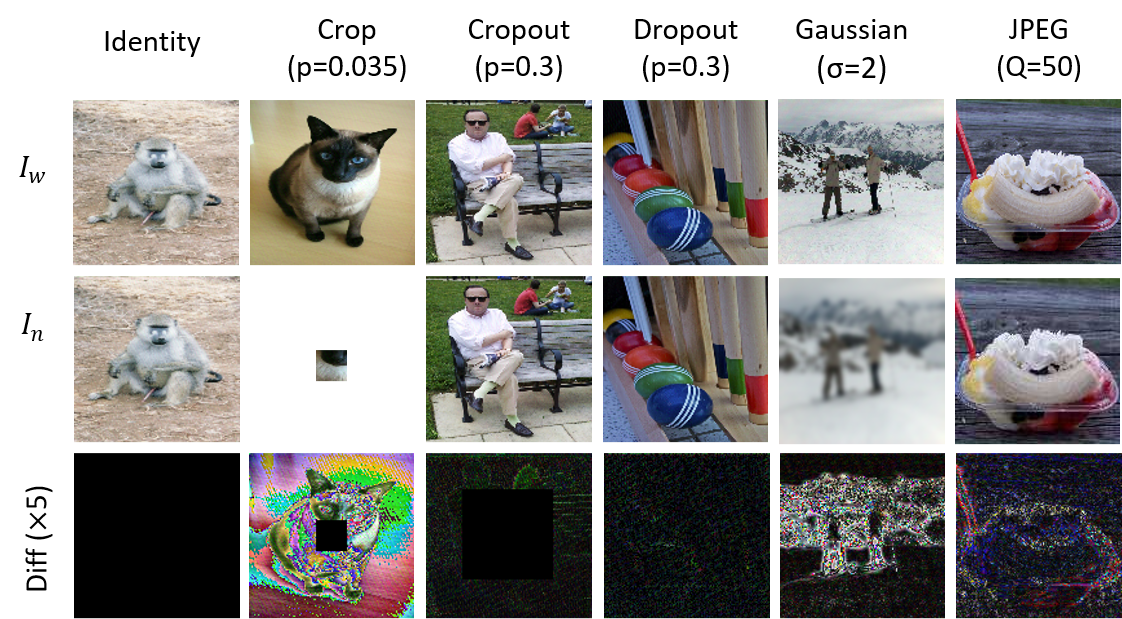}
    \caption{Illustration of different noises. The first row is the watermarked image $I_w$, the second row is the noised image $I_n$. The third row is the magnified difference with $| I_w - I_n| \times 5$. }
    \label{noise_type}
\end{figure}

\begin{table*}[htbp]
\centering
    \caption{\label{psnr} Objective comparison for HiDDeN \cite{zhu2018hidden} and our model trained with different noise layers. Here we list the average PSNR metric between the watermarked image $I_w$ and the cover image $I_c$. The last column refers to the combined model for all noises, and the rest columns refer to specialized models for specific noises.
    }
\begin{tabular}{c|cccccc|c }
\hline
 & \multirow{2}{*}{Identity} & Crop & Cropout & Dropout  & Gaussian  & JPEG & \multirow{2}{*}{Combined}  \\
  & & (p=0.035) &(p=0.3)&(p=0.3)  & ($\sigma$=2)& (Q=50) & \\
\hline
{HiDDeN \cite{zhu2018hidden}} & 36.74 & 32.70 & 31.94 & \textbf{34.39} & 30.38 & 32.64 & 32.92 \\
{Ours}           & \textbf{37.88}  & \textbf{34.30}  & \textbf{32.26}  & 30.31  & \textbf{37.03}  & \textbf{36.16}  &   \textbf{32.99}   \\

\hline
\end{tabular}
\end{table*}

\begin{figure*}[htb]
    \centering
    \includegraphics[width=1.0\linewidth]{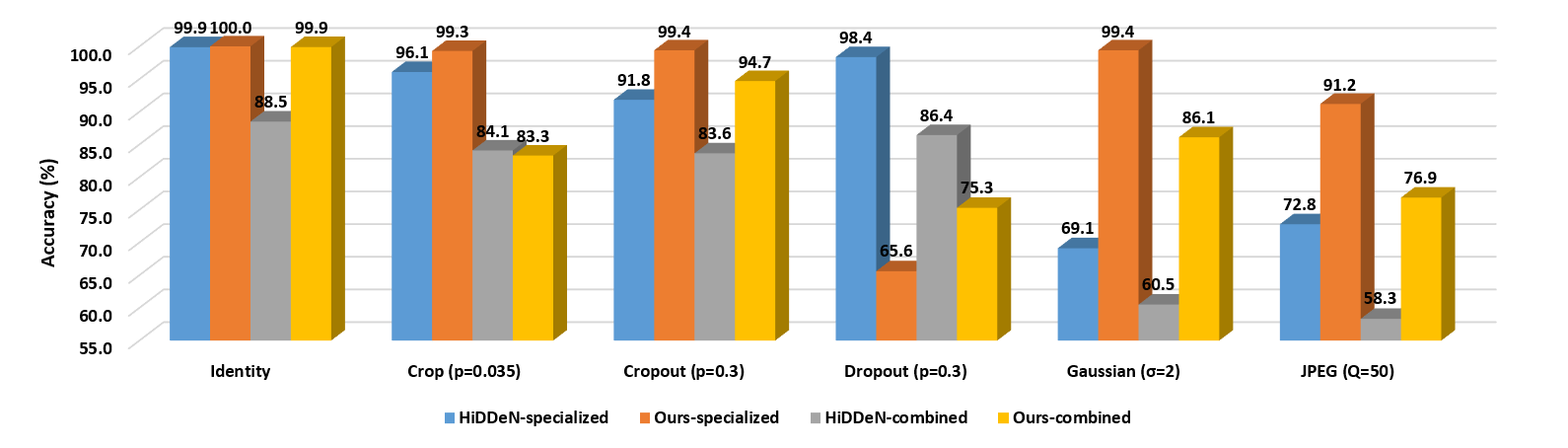}
    \caption{Robustness comparison against different distortions. Each cluster corresponds to a special distortion. 'combined' refers to training and testing on all 6 types of distortion, and 'specialized' means training and testing on the specific distortion type.}
    \label{bit_accuracy}
\end{figure*}

\begin{figure*}[htb]
    \centering
    \includegraphics[width=1.0\linewidth]{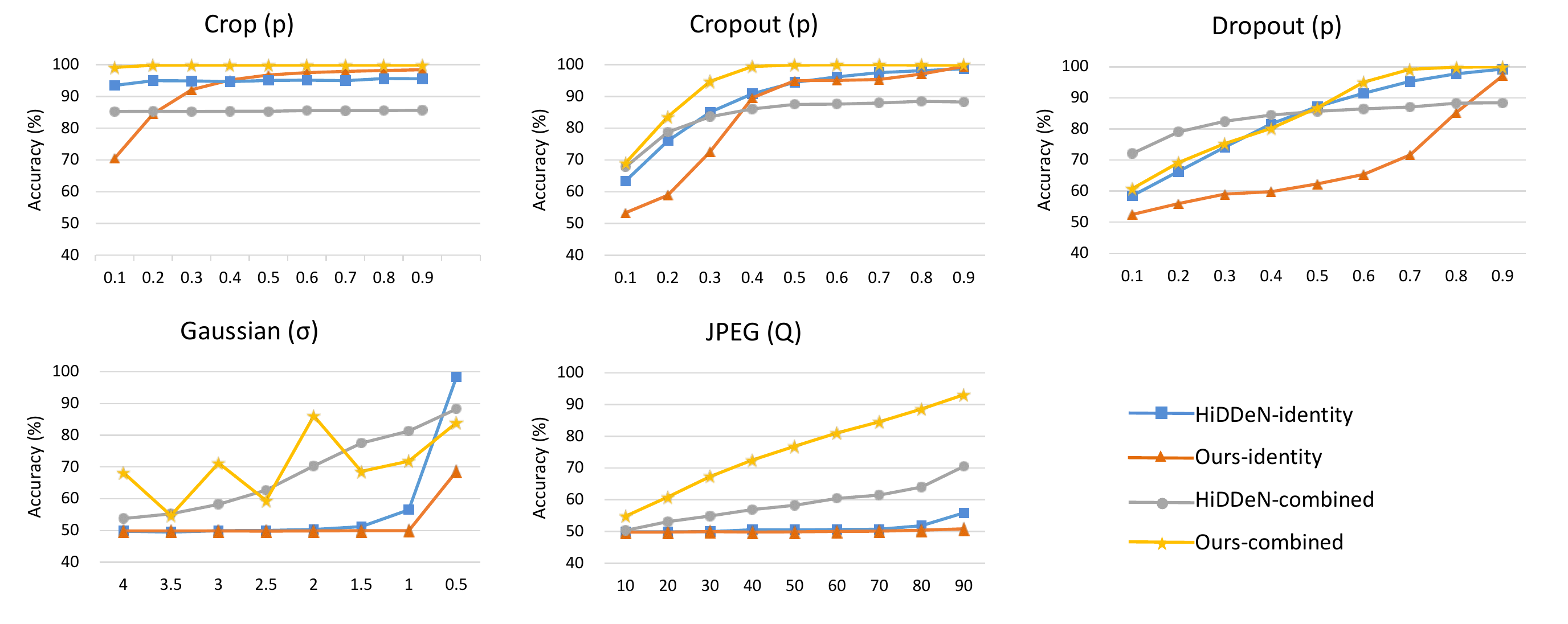}
    \caption{The accuracy of bit message recovery under five common distortions with various intensities. Here the compared Identity model is trained without noise, while the combined model is trained on all distortion types.}
    \label{robustness}
\end{figure*}

\begin{figure*}[htb]
    \centering
    \includegraphics[width=.95\linewidth]{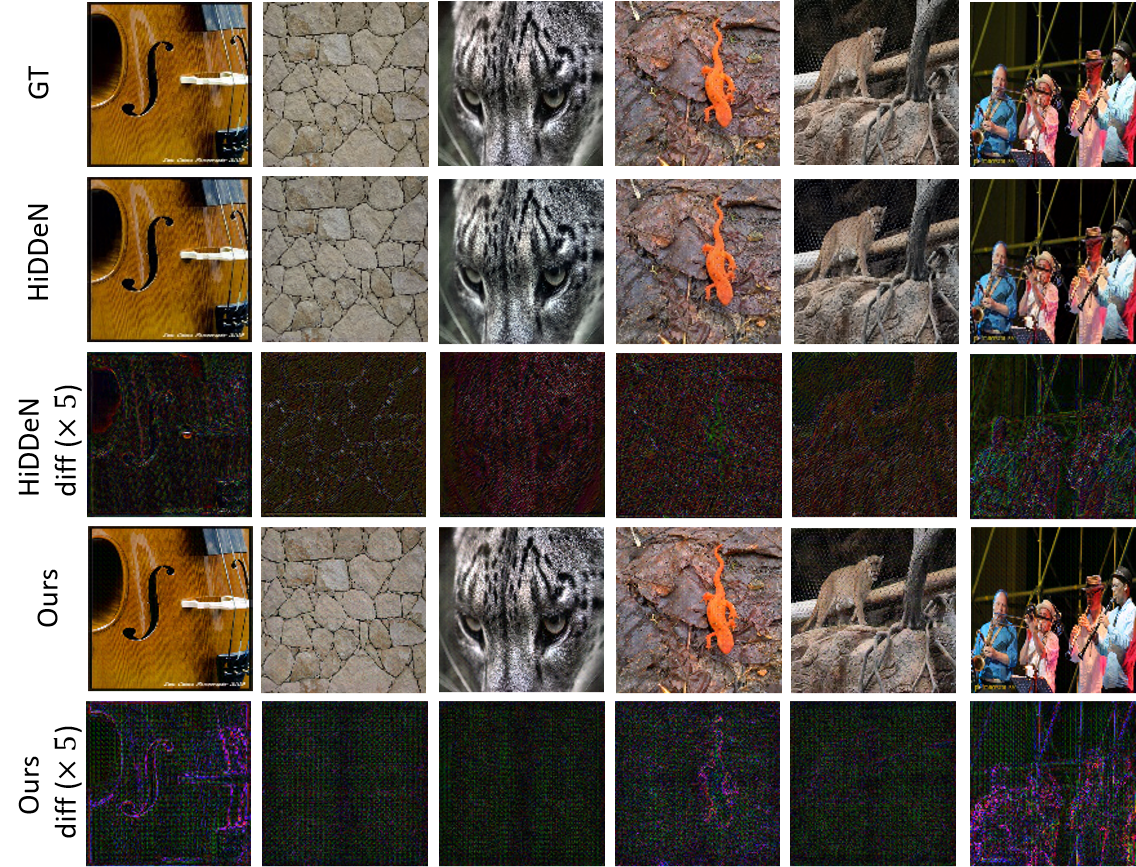}
    \caption{Visual comparison on some watermarked images. The first row is the cover image, regarded as ground-truth (GT). The second and the fourth row are the watermark images generated by HiDDeN~\cite{zhu2018hidden} and our method, respectively. The third and the fifth row are the difference between the watermark image and its GT, which is $\times 5$ magnified for visualization. }
    \label{visual_comparison}
\end{figure*}

In practice, watermarked images would suffer from various distortions during compression, transmission and interactive editing operations.
The robustness against these different attacks (or noises), which may destroy the embedded watermarks in the real world, is one of the important issues for a digital image watermarking algorithm.
In order to improve the robustness, here we introduce the noise layer $N$ to simulate various image distortions, including Identity, Crop, Cropout, Dropout, Gaussian and JPEG operations. 
See \figref{noise_type} for more details and examples. 
Specifically, as a component of IWN, we expect the proposed noise layer is also differentiable so that the whole network can be trained in an end-to-end style. 
Next, we will detailed discuss these different distortions in terms of whether they are differentiable or not.

\textbf{Differentiable Noises.}
Most watermarking noises, including Identity, Crop, Cropout, Dropout, and Gaussian, are inherently differentiable, so we add them to our framework directly.
For the Identity noise, we do not change the watermarked image at all.
Crop refers to producing a $H' \times W'$ rectangle by randomly cropping from the $H \times W$ watermarked image, and the calculated percentage $ p = \frac{H' \times W'}{H \times W}$ is introduced to control the remaining ratio of the watermarked image. 
Cropout means randomly replacing the $H' \times W'$ rectangle of the $H \times W$ cover image with counterpart of the watermarked image, and similarly, the ratio $p = \frac{H' \times W'}{H \times W}$.
Dropout also replaces some cover image pixels with the watermarked image pixels similar to Cropout, while the difference is that instead of replacing a whole area, the former randomly selects some pixels for replacement based on the remaining ratio $p$.
Finally, Gaussian blurs the watermarked image with a gaussian kernel of the given width $\sigma$.

\textbf{Quantization.} 
Watermarked images are being widely applied in storage and transmission, so they must be converted into commonly used image formats, such as the 8-bit RGB format (\ie 8 bits for each color channel). 
In the practical implementation, we need a differentiable quantization module to convert the floating-point values of the INN's outputs to 8-bit unsigned integers. 
To ensure the gradients back propagation during training, here the rounding operation is used as the quantization module, and the Straight-Through Estimator~\cite{INN_2019IRB} is adopted when calculating the gradients.
In our solution, we combine this quantization noise in all our training and testing experiments, such that our watermarking system could effectively deal with quantization error. 

\textbf{JPEG compression.}
Similar to the quantization operator, the noise produced by JPEG compression is non-differentiable due to the quantization step in the compression framework. 
To solve the gradients back propagation problem during training, 
we follow~\cite{shin2017jpeg} to simulate the quantization step in the standard JPEG compression through the following equation, 
\begin{equation}
    \label{eq:jpeg}
    \lfloor x \rceil_{approx} = \lfloor x \rceil + (x - \lfloor x \rceil)^3 ,
\end{equation}
where $\lfloor \cdot \rceil$ is the rounding function and $\lfloor \cdot \rceil_{approx}$ is the differentiable approximation of the rounding function which has non-zero derivatives nearly everywhere.
Note that in our solution we use the real JPEG compression instead of constructing a JPEG simulator during testing.
By transforming the non-differentiable part into an approximate representation that is derivable, we construct a completely differentiable noise layer, with which our IWN can be efficiently trained in an end-to-end way.

\section{Experiments}  \label{sec_experiments}
\subsection{Experimental Setup}
\label{sec:Setup}
We implement our IWN with PyTorch and train our model with the Adam Optimizer. The learning rate is set to 2e-4 and the batch size is 6.
We use 16 invertible blocks for embedding a $l = 30$ bits message and the message is divided into $c = 10$ groups.
In addition, all elements of constant matrix $Z$ are set to 0.5.
Our IWN is trained on DIV2K \cite{div2k_2017_CVPR_Workshops} and Flickr2K \cite{flick2k_2017_CVPR_Workshops} datasets, and it is tested on a subset of ImageNet \cite{imagenet_ijcv15} which contains 1000 images.
All training images are cropped into 480$\times$480 patches and resized to 128$\times$128 during training, while the test images are all resized to 128$\times$128.
Flipping and rotation are randomly used for data augmentation.
The quality factor Q of the JPEG simulator is uniformly sampled from \{50, 60, 70, 80, 90\}.

In our system, the loss weights are specified as $\omega_{m} = 1.0$, $\omega_{z} = 1.0$, and $\omega_{c} = 0.1$.
Besides, the weight of $\omega_{w}$ varies according to different training stages.
For instance, when the noise layer is with the Identity layer, which means that no distortions are applied on the watermarked image, $\omega_{w}$ is set to 32.
In other cases, $\omega_{w}$ is first set to 0.1 until the system is converged, and then $\omega_{w}$ is refined as 48.0 until convergence.
In other words, we first train the robustness against various distortions when extracting the watermark, and then train the imperceptibility of our watermarking solution.
All experiments are conducted on two Nvidia RTX 2080Ti GPUs.
\begin{table*}[htbp]
\centering
    \caption{\label{different_blocks} Ablation experiments for the number of the invertible blocks. We list the PSNR of the watermark images (the second column) and the accuracy of bit message recovery under 6 distortions (the last 6 columns).}
\begin{tabular}{c|c|cccccc}
\hline
\multirow{2}{*}{Block number}& PSNR & \multirow{2}{*}{Identity}  & Crop & Cropout  & Dropout  & Gaussian  & JPEG  \\
            & (dB) &           & (p=0.035)&(p=0.3)&(p=0.3)  & ($\sigma$=2)& (Q=50) \\
\hline
 4    &  30.80  & 0.8621  & 0.7987  & 0.8181  & 0.6428  &  0.7523 & 0.5433  \\
 8    &  30.21  & 0.9762  & 0.8187  & 0.9143  & \underline{0.7329}  &  0.8449  & 0.6306 \\
 12   &  \underline{30.97}  & \underline{0.9949}  &\textbf{ 0.8594}  & \textbf{0.9655}  & 0.7138  &  \textbf{0.9364} & \underline{0.6604} \\
 16   &  \textbf{32.99}  & \textbf{0.9994}  & \underline{0.8331}  & \underline{0.9471}  & \textbf{0.7529}  &  \underline{0.8611} & \textbf{0.7687} \\
\hline
\end{tabular}
\end{table*}

\begin{table*}[htbp]
\centering
    \caption{\label{different_length} Ablation experiments for the proposed bit message normalization module. 
    }
\begin{tabular}{c|c|cccccc}
\hline

\multirow{2}{*}{$l$ \& $c$}& PSNR & \multirow{2}{*}{Identity}  & Crop & Cropout  & Dropout  & Gaussian  & JPEG  \\
       & (dB) &           & (p=0.035)&(p=0.3)&(p=0.3)  & ($\sigma$=2)& (Q=50) \\
\hline
 30 \& 10        & \textbf{32.99}  & \textbf{0.9994}  & \textbf{0.8331}  & \textbf{0.9471}  & \textbf{0.7529}  &  \textbf{0.8611} & \textbf{0.7687}  \\
 40  \& 10       & 31.34  & 0.9187  & 0.7055  & 0.8211 & 0.6297  & 0.8087  & 0.7313 \\
 50  \& 10       & 31.96  & 0.7494  & 0.6708  & 0.7342 & 0.6129 & 0.6879  & 0.6585 \\
 60  \& 10       & 30.50  & 0.7212  & 0.6275  & 0.6749 & 0.5913  & 0.6256  & 0.6403 \\
 \hline
  30 \& 30        & 30.36  & 0.7470  & 0.7271  & 0.7415 & 0.6306  & 0.7455  & 0.5788 \\
\hline
\end{tabular}
\end{table*}

\subsection{Metrics}
We evaluate our method mainly on robustness and imperceptibility which are more important than capacity for watermarking algorithm generally.
Specifically, we measure the imperceptibility using peak signal-to-noise ratio (PSNR) between the cover image and watermarked image.
And we measure robustness using bit accuracy, which is the percentage of identical bits between the original message $M$ and the extracted message $\hat{M}$ to total bits of the message.

\subsection{Comparison}

We take HiDDeN~\cite{zhu2018hidden} as the baseline method for comparison since it is a well studied model and a commonly used benchmark.
%
%
%
We reconduct the experiments of HiDDeN~\cite{zhu2018hidden} with the open source code.
%
%
%
Following HiDDeN~\cite{zhu2018hidden}, watermarked images are exposed to the following 6 distortions: Identity, Crop, Cropout, Dropout, Gaussian, and JPEG compression.
We respectively control the intensity of distortions with the following scalars: the remaining ratio $p$ for Crop, Cropout and Dropout, the kernel width $\sigma$ for Gaussian, and the quality factor Q for JPEG compression.
These scalars are identical with those adopted in HiDDeN~\cite{zhu2018hidden} during testing.
Specialized models are optimized to be resistant to specific distortions aforementioned, and the final combined model is trained to be robust against all kinds of distortions.
In order to further improve the robustness against JPEG compression from the real world, the noise layer is randomly sampled from set \{Identity, Crop, Cropout, Dropout, Gaussian, JPEG\} with a probability distribution of \{0.05, 0.05, 0.1, 0.15, 0.65\} for each mini-batch during training the combined model.
We report both the bit accuracy and PSNR when various distortions are applied on watermarked images.
It is worth noticing that these two metrics may present conflicting evaluation results.
For instance, a higher PSNR value usually means that the embedded message changes less information to the cover image, which makes it more difficult to accurately recover the bit sequence message, and the corresponding bit accuracy would decrease. 
For a well-designed watermarking algorithm, it should not be with a high PSNR but low bit accuracy or high bit accuracy but low PSNR.
%
%
Therefore, we deliberately avoid this conflicting situation for the fair comparison.

\subsubsection{Quantitative Results}
In \tabref{psnr}, it shows the PSNR metric between cover images $I_c$ and watermarked images $I_w$ produced by 6 specialized models against the corresponding noises and 1 combined model.
Moreover, \figref{bit_accuracy} illustrates the bit accuracy of different models against 6 distortions.
In general, when compared with those specialized models, our 5 models, \ie Identity, Crop, Cropout, Gaussian, and JPEG, have higher PSNR values.
Meanwhile, our solution obtains higher bit accuracy than the baseline method.
Especially, our method achieves both +3.52 dB gains than the baseline for the imperceptibility under JPEG compression, and 18.4\% higher bit accuracy in terms of robustness.
When comparing the combined model, we have almost the same performance for the PSNR metric, but the robustness of our algorithm is much better than the baseline against the Identity, Cropout, Gaussian, and JPEG compression distortions, among which 18.6\% higher bit accuracy is achieved under the JPEG compression distortion.
Besides, these two metrics, \ie the PSNR and the bit accuracy, demonstrate that our method achieves a better balance than that of HiDDeN~\cite{zhu2018hidden} between the imperceptibility and the robustness of watermarking.
%
%

\figref{robustness} provides a more comprehensive comparison in the bit accuracy against various intensities of distortions. 
%
%
In general, our combined model achieves better performance than HiDDeN~\cite{zhu2018hidden} for resisting distortions.
%
%
Although our method fails in Dropout (p=0.3) according to \figref{bit_accuracy}, the curve of Dropout in \figref{robustness} shows our method surpasses HiDDeN~\cite{zhu2018hidden} when the remaining ratio $p>0.5$.
When comparing the identity model to combined model, we can see that the noise layer $N$ has obvious benefits for Gaussian and JPEG compression distortions, as the identity model generally fails under such two attacks.  
%
%
%
%

\subsubsection{Qualitative Results}
\figref{visual_comparison} provides visual comparison of watermarked images produced by combined models. 
Both the watermark images and the corresponding $\times 5$ magnified differences compared to the original images demonstrate that the bit message is successfully embedded in the images in an imperceptible way.
The watermark images generated by our model and HiDDeN~\cite{zhu2018hidden} look very similar to the corresponding cover images, which is consistent with the PSNR metric reported in \tabref{psnr}. 

\subsection{Ablation Study}
Here we discuss how the hyper-parameters, the number of invertible blocks and the length of the bit message affect the performance of our method.
Because our goal is to get a robust watermarking model, the experiments in this part are conducted on the combined model rather than any specialized one.

Firstly, we discuss the number of the invertible blocks in our INN module, and we report the performance in \tabref{different_blocks}.
In general, our solution gets better performance when with more blocks.
It is reasonable to understand, as more invertible blocks imply more trainable parameters.
It is particularly worth noting that when the number of blocks changes from 12 to 16, the bit accuracy under JPEG compression significantly increases by 10\% (see the last two rows of the last column of \tabref{different_blocks}), and the PSNR of the watermarked images has also increased by 2.02 dB.  
The bottleneck of our method is the robustness against JPEG compression, as when using 8 or 12 blocks, the bit accuracy is above 0.7 over other distortions except for JPEG compression.
On the other hand, although the model of 16 invertible blocks does not always bring the best performance, it achieves the highest PSNR values and the best robustness against the JPEG compression. Thus, our final model chooses 16 invertible blocks.

Secondly, we verify the effectiveness of our bit message normalization module. 
To this end, we carry out different experiments by changing two variables, \ie, the bit length of the embedded message $l$, and the number $c$ for dividing the message into groups.
The detailed experimental results are shown in \tabref{different_length}, which includes two kind of well-trained models with $c$ as 10 and 30, respectively.
To study the performance of our model fluctuated with different bit message length $l$, we test the models with $c=10$ message groups when $l$ is set to 30, 40, 50, and 60, \ie each channel represents 3, 4, 5 and 6 bits, respectively.
The other experimental settings are the same as we mentioned in \secref{sec:Setup}.  
Obviously, the bit accuracy decreases when the message becomes longer. 
This follows the conflict between the capacity and robustness of watermarking algorithm. 
We further carry out an experiment without the bit message normalization module.
Specifically, we directly treat the binary bit message as float numbers (0.0 or 1.0) like HiDDeN~\cite{zhu2018hidden}, and both $l$ and $c$ are thus set to 30. 
Without our bit message normalization, all the evaluation metrics drop dramatically, as shown in the last row of \tabref{different_length}.

\subsection{Limitations and Future Work}
\begin{figure}[t]
    \centering
    \includegraphics[width=.95\linewidth]{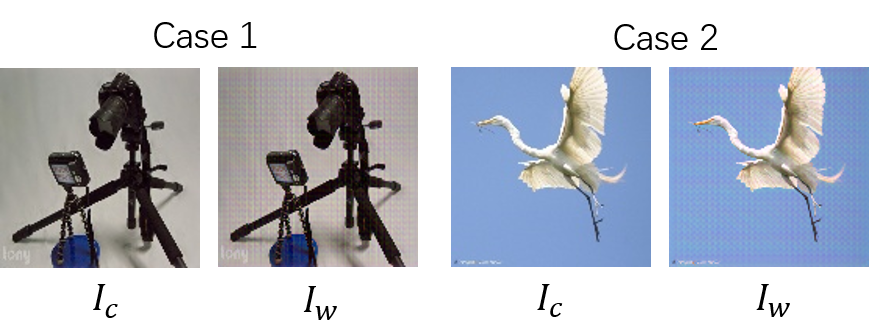}
    \caption{Limitations of our solution. The watermarked images $I_w$ produced by our method may contain some visual artifacts in the smooth region. }
    \label{limitation}
\end{figure}

In order to resist the practical distortions especially introduced by JPEG compression, our watermarked images $I_w$ may generate some visual artifacts when there exist many smooth regions. 
In \figref{limitation}, the background (case 1) and the sky (case 2) presents this phenomenon, respectively. 
Embedding information in a smooth area is inherently difficult till now.
This issue may be improved through embedding the watermark by paying more attention to the edges or rich texture areas of the image.
Besides, introducing some smoothing loss for the watermarked image during training, such as the Fourier transform loss and structural similarity index measure (SSIM) loss, may also be helpful to alleviate this problem.
We will further explore how to remove these visual artifacts in the future.

\section{Conclusion}  \label{sec_conclusion}

In this paper, we have presented an invertible watermarking network (IWN) for robust blind digital image watermarking. 
Our compact IWN utilizes the invertible neural network (INN) to embed and extract the watermark message with an end-to-end training style.
To promote the watermarking robustness against various practical distortions, we specifically introduce a noise layer to simulate various attacks.
Moreover, we propose a simple but effective bit message normalization module to further enhance the watermarking robustness.
Extensive experiments demonstrate the superiority of our method against the commonly used baseline.
In the future, we will also explore the application of our framework in cross-media channels, such as printing and photographing, screen photographing, and audio-visual watermarking in other multimedia domains.
%



\bibliographystyle{IEEEtran}
\bibliography{ref}

\begin{thebibliography}{10}
\providecommand{\url}[1]{#1}
\csname url@samestyle\endcsname
\providecommand{\newblock}{\relax}
\providecommand{\bibinfo}[2]{#2}
\providecommand{\BIBentrySTDinterwordspacing}{\spaceskip=0pt\relax}
\providecommand{\BIBentryALTinterwordstretchfactor}{4}
\providecommand{\BIBentryALTinterwordspacing}{\spaceskip=\fontdimen2\font plus
\BIBentryALTinterwordstretchfactor\fontdimen3\font minus
  \fontdimen4\font\relax}
\providecommand{\BIBforeignlanguage}[2]{{%
\expandafter\ifx\csname l@#1\endcsname\relax
\typeout{** WARNING: IEEEtran.bst: No hyphenation pattern has been}%
\typeout{** loaded for the language `#1'. Using the pattern for}%
\typeout{** the default language instead.}%
\else
\language=\csname l@#1\endcsname
\fi
#2}}
\providecommand{\BIBdecl}{\relax}
\BIBdecl

\bibitem{chen2001quantization}
B.~Chen and G.~W. Wornell, ``Quantization index modulation: A class of provably
  good methods for digital watermarking and information embedding,''
  \emph{{IEEE} Trans. Inf. Theory}, vol.~47, no.~4, pp. 1423--1443, 2001.

\bibitem{faundez2007speaker}
M.~Faundez-Zanuy, M.~Hagm{\"u}ller, and G.~Kubin, ``Speaker identification
  security improvement by means of speech watermarking,'' \emph{Pattern
  Recognition}, vol.~40, no.~11, pp. 3027--3034, 2007.

\bibitem{broughton1989interactive}
R.~S. Broughton and W.~C. Laumeister, ``Interactive video method and
  apparatus,'' Feb.~21 1989, uS Patent 4,807,031.

\bibitem{loan2018secure}
N.~A. Loan, N.~N. Hurrah, S.~A. Parah, J.~W. Lee, J.~A. Sheikh, and G.~M. Bhat,
  ``Secure and robust digital image watermarking using coefficient differencing
  and chaotic encryption,'' \emph{IEEE Access}, vol.~6, pp. 19\,876--19\,897,
  2018.

\bibitem{Pexaras2019OptimizationAH}
K.~Pexaras, I.~G. Karybali, and E.~Kalligeros, ``Optimization and hardware
  implementation of image and video watermarking for low-cost applications,''
  \emph{IEEE Trans. Circuits Syst. {I}}, vol.~66, pp. 2088--2101, 2019.

\bibitem{tao2014robust}
H.~Tao, L.~Chongmin, J.~M. Zain, and A.~N. Abdalla, ``Robust image watermarking
  theories and techniques: A review,'' \emph{Journal of applied research and
  technology}, vol.~12, no.~1, pp. 122--138, 2014.

\bibitem{zhu2018hidden}
J.~Zhu, R.~Kaplan, J.~Johnson, and L.~Fei-Fei, ``Hidden: Hiding data with deep
  networks,'' in \emph{Eur. Conf. Comput. Vis.}, 2018, pp. 657--672.

\bibitem{tancik2020stegastamp}
M.~Tancik, B.~Mildenhall, and R.~Ng, ``Stegastamp: Invisible hyperlinks in
  physical photographs,'' in \emph{IEEE Conf. Comput. Vis. Pattern Recog.},
  2020, pp. 2117--2126.

\bibitem{luo2020distortion}
X.~Luo, R.~Zhan, H.~Chang, F.~Yang, and P.~Milanfar, ``Distortion agnostic deep
  watermarking,'' in \emph{IEEE Conf. Comput. Vis. Pattern Recog.}, 2020, pp.
  13\,548--13\,557.

\bibitem{ray2020recent}
A.~Ray and S.~Roy, ``Recent trends in image watermarking techniques for
  copyright protection: a survey,'' \emph{International Journal of Multimedia
  Information Retrieval}, pp. 1--22, 2020.

\bibitem{byrnes2021data}
O.~Byrnes, W.~La, H.~Wang, C.~Ma, M.~Xue, and Q.~Wu, ``Data hiding with deep
  learning: A survey unifying digital watermarking and steganography,''
  \emph{arXiv preprint arXiv:2107.09287}, 2021.

\bibitem{KANDI2017247}
H.~Kandi, D.~Mishra, and S.~R.~S. Gorthi, ``Exploring the learning capabilities
  of convolutional neural networks for robust image watermarking,''
  \emph{Computers \& Security}, vol.~65, pp. 247--268, 2017.

\bibitem{Wen2019ROMarkAR}
B.~Wen and S.~Ayd{\"o}re, ``Romark: A robust watermarking system using
  adversarial training,'' \emph{ArXiv}, vol. abs/1910.01221, 2019.

\bibitem{wengrowski2019light}
E.~Wengrowski and K.~Dana, ``Light field messaging with deep photographic
  steganography,'' in \emph{IEEE Conf. Comput. Vis. Pattern Recog.}, 2019, pp.
  1515--1524.

\bibitem{jia2020rihoop}
J.~Jia, Z.~Gao, K.~Chen, M.~Hu, X.~Min, G.~Zhai, and X.~Yang, ``Rihoop: Robust
  invisible hyperlinks in offline and online photographs,'' \emph{IEEE Trans.
  Cybern.}, 2020.

\bibitem{yu2020attention}
C.~Yu, ``Attention based data hiding with generative adversarial networks,'' in
  \emph{AAAI}, vol.~34, no.~01, 2020, pp. 1120--1128.

\bibitem{zhong2020automated}
X.~Zhong, P.~C. Huang, S.~Mastorakis, and F.~Y. Shih, ``An automated and robust
  image watermarking scheme based on deep neural networks,'' \emph{IEEE Trans.
  Multimedia}, 2020.

\bibitem{cheng2021iicnet}
K.~L. Cheng, Y.~Xie, and Q.~Chen, ``Iicnet: A generic framework for reversible
  image conversion,'' in \emph{Int. Conf. Comput. Vis.}, 2021, pp. 1991--2000.

\bibitem{lu2021large}
S.-P. Lu, R.~Wang, T.~Zhong, and P.~L. Rosin, ``Large-capacity image
  steganography based on invertible neural networks,'' in \emph{IEEE Conf.
  Comput. Vis. Pattern Recog.}, 2021, pp. 10\,816--10\,825.

\bibitem{INN_2018analyzing}
L.~Ardizzone, J.~Kruse, C.~Rother, and U.~K{\"o}the, ``Analyzing inverse
  problems with invertible neural networks,'' in \emph{Int. Conf. Learn.
  Represent.}, 2018.

\bibitem{van1994digital}
R.~G. Van~Schyndel, A.~Z. Tirkel, and C.~F. Osborne, ``A digital watermark,''
  in \emph{Proceedings of 1st international conference on image processing},
  vol.~2.\hskip 1em plus 0.5em minus 0.4em\relax IEEE, 1994, pp. 86--90.

\bibitem{katzenbeisser2000digital}
S.~Katzenbeisser and F.~Petitcolas, ``Digital watermarking,'' \emph{Artech
  House, London}, vol.~2, 2000.

\bibitem{cox2002digital}
I.~J. Cox, M.~L. Miller, J.~A. Bloom, and C.~Honsinger, \emph{Digital
  watermarking}.\hskip 1em plus 0.5em minus 0.4em\relax Springer, 2002,
  vol.~53.

\bibitem{cox2007digital}
I.~Cox, M.~Miller, J.~Bloom, J.~Fridrich, and T.~Kalker, \emph{Digital
  watermarking and steganography}.\hskip 1em plus 0.5em minus 0.4em\relax
  Morgan kaufmann, 2007.

\bibitem{haddad2020joint}
S.~Haddad, G.~Coatrieux, A.~Moreau-Gaudry, and M.~Cozic, ``Joint
  watermarking-encryption-jpeg-ls for medical image reliability control in
  encrypted and compressed domains,'' \emph{{IEEE} Trans. Inf. Forensics
  Security.}, vol.~15, pp. 2556--2569, 2020.

\bibitem{asikuzzaman2014imperceptible}
M.~Asikuzzaman, M.~J. Alam, A.~J. Lambert, and M.~R. Pickering, ``Imperceptible
  and robust blind video watermarking using chrominance embedding: a set of
  approaches in the dt cwt domain,'' \emph{{IEEE} Trans. Inf. Forensics
  Security.}, vol.~9, no.~9, pp. 1502--1517, 2014.

\bibitem{ma2019xmark}
H.~Ma, C.~Jia, S.~Li, W.~Zheng, and D.~Wu, ``Xmark: dynamic software
  watermarking using collatz conjecture,'' \emph{{IEEE} Trans. Inf. Forensics
  Security.}, vol.~14, no.~11, pp. 2859--2874, 2019.

\bibitem{gao2021thermotag}
Y.~Gao, W.~Wang, Y.~Jin, C.~Zhou, W.~Xu, and Z.~Jin, ``Thermotag: A hidden id
  of 3d printers for fingerprinting and watermarking,'' \emph{{IEEE} Trans.
  Inf. Forensics Security.}, vol.~16, pp. 2805--2820, 2021.

\bibitem{hou2017blind}
J.-U. Hou, D.-G. Kim, and H.-K. Lee, ``Blind 3d mesh watermarking for 3d
  printed model by analyzing layering artifact,'' \emph{{IEEE} Trans. Inf.
  Forensics Security.}, vol.~12, no.~11, pp. 2712--2725, 2017.

\bibitem{liu2018patchwork}
Z.~Liu, Y.~Huang, and J.~Huang, ``Patchwork-based audio watermarking robust
  against de-synchronization and recapturing attacks,'' \emph{{IEEE} Trans.
  Inf. Forensics Security.}, vol.~14, no.~5, pp. 1171--1180, 2018.

\bibitem{wu2020watermarking}
H.~Wu, G.~Liu, Y.~Yao, and X.~Zhang, ``Watermarking neural networks with
  watermarked images,'' \emph{{IEEE} Trans. Circuits Syst. Video Technol.},
  2020.

\bibitem{su2018robust}
Q.~Su and B.~Chen, ``Robust color image watermarking technique in the spatial
  domain,'' \emph{Soft Computing}, vol.~22, no.~1, pp. 91--106, 2018.

\bibitem{li2013general}
X.~Li, B.~Li, B.~Yang, and T.~Zeng, ``General framework to
  histogram-shifting-based reversible data hiding,'' \emph{IEEE Trans. Image
  Process.}, vol.~22, no.~6, pp. 2181--2191, 2013.

\bibitem{deng2010local}
C.~Deng, X.~Gao, X.~Li, and D.~Tao, ``Local histogram based geometric invariant
  image watermarking,'' \emph{Signal Processing}, vol.~90, no.~12, pp.
  3256--3264, 2010.

\bibitem{2013dabas}
P.~Dabas and K.~Khanna, ``A study on spatial and transform domain watermarking
  techniques,'' \emph{International Journal of Computer Applications}, vol.~71,
  pp. 38--41, 06 2013.

\bibitem{2018kumar}
C.~Kumar, A.~K. Singh, and P.~Kumar, ``A recent survey on image watermarking
  techniques and its application in e-governance,'' \emph{Multimedia Tools
  Appl.}, vol.~77, no.~3, pp. 3597--3622, 2018.

\bibitem{huang2019enhancing}
Y.~Huang, B.~Niu, H.~Guan, and S.~Zhang, ``Enhancing image watermarking with
  adaptive embedding parameter and psnr guarantee,'' \emph{IEEE Trans.
  Multimedia}, vol.~21, no.~10, pp. 2447--2460, 2019.

\bibitem{hamidi2018hybrid}
M.~Hamidi, M.~El~Haziti, H.~Cherifi, and M.~El~Hassouni, ``Hybrid blind robust
  image watermarking technique based on dft-dct and arnold transform,''
  \emph{Multimedia Tools Appl.}, vol.~77, no.~20, pp. 27\,181--27\,214, 2018.

\bibitem{guo2002digital}
H.~Guo and N.~D. Georganas, ``Digital image watermarking for joint ownership,''
  in \emph{ACM Int. Conf. Multimedia}, 2002, pp. 362--371.

\bibitem{bao2005image}
P.~Bao and X.~Ma, ``Image adaptive watermarking using wavelet domain singular
  value decomposition,'' \emph{{IEEE} Trans. Circuits Syst. Video Technol.},
  vol.~15, no.~1, pp. 96--102, 2005.

\bibitem{bi2007robust}
N.~Bi, Q.~Sun, D.~Huang, Z.~Yang, and J.~Huang, ``Robust image watermarking
  based on multiband wavelets and empirical mode decomposition,'' \emph{IEEE
  Trans. Image Process.}, vol.~16, no.~8, pp. 1956--1966, 2007.

\bibitem{kang2003dwt}
X.~Kang, J.~Huang, Y.~Q. Shi, and Y.~Lin, ``A dwt-dft composite watermarking
  scheme robust to both affine transform and jpeg compression,'' \emph{{IEEE}
  Trans. Circuits Syst. Video Technol.}, vol.~13, no.~8, pp. 776--786, 2003.

\bibitem{sadreazami2018robust}
H.~Sadreazami and M.~Amini, ``A robust image watermarking scheme using local
  statistical distribution in the contourlet domain,'' \emph{IEEE Trans.
  Circuits Syst. {II}}, vol.~66, no.~1, pp. 151--155, 2018.

\bibitem{zhang2011affine}
H.~Zhang, H.~Shu, G.~Coatrieux, J.~Zhu, Q.~J. Wu, Y.~Zhang, H.~Zhu, and L.~Luo,
  ``Affine legendre moment invariants for image watermarking robust to
  geometric distortions,'' \emph{IEEE Trans. Image Process.}, vol.~20, no.~8,
  pp. 2189--2199, 2011.

\bibitem{pereira2000robust}
S.~Pereira and T.~Pun, ``Robust template matching for affine resistant image
  watermarks,'' \emph{IEEE Trans. Image Process.}, vol.~9, no.~6, pp.
  1123--1129, 2000.

\bibitem{xiang2008invariant}
S.~Xiang, H.~J. Kim, and J.~Huang, ``Invariant image watermarking based on
  statistical features in the low-frequency domain,'' \emph{{IEEE} Trans.
  Circuits Syst. Video Technol.}, vol.~18, no.~6, pp. 777--790, 2008.

\bibitem{tian2013ldft}
H.~Tian, Y.~Zhao, R.~Ni, L.~Qin, and X.~Li, ``Ldft-based watermarking resilient
  to local desynchronization attacks,'' \emph{IEEE Trans. Cybern.}, vol.~43,
  no.~6, pp. 2190--2201, 2013.

\bibitem{mun2017robust}
S.-M. Mun, S.-H. Nam, H.-U. Jang, D.~Kim, and H.-K. Lee, ``A robust blind
  watermarking using convolutional neural network,'' \emph{arXiv preprint
  arXiv:1704.03248}, 2017.

\bibitem{ahmadi2020redmark}
M.~Ahmadi, A.~Norouzi, N.~Karimi, S.~Samavi, and A.~Emami, ``Redmark: Framework
  for residual diffusion watermarking based on deep networks,'' \emph{Expert
  Systems with Applications}, vol. 146, p. 113157, 2020.

\bibitem{zhang2020udh}
C.~Zhang, P.~Benz, A.~Karjauv, G.~Sun, and I.~S. Kweon, ``Udh: Universal deep
  hiding for steganography, watermarking, and light field messaging,''
  \emph{Adv. Neural Inform. Process. Syst.}, vol.~33, pp. 10\,223--10\,234,
  2020.

\bibitem{liu2019novel}
Y.~Liu, M.~Guo, J.~Zhang, Y.~Zhu, and X.~Xie, ``A novel two-stage separable
  deep learning framework for practical blind watermarking,'' in \emph{ACM Int.
  Conf. Multimedia}, 2019, pp. 1509--1517.

\bibitem{est_2014nice}
L.~Dinh, D.~Krueger, and Y.~Bengio, ``{NICE}: Non-linear independent components
  estimation,'' \emph{arXiv preprint arXiv:1410.8516}, 2014.

\bibitem{est_2016nvp}
L.~Dinh, J.~Sohl-Dickstein, and S.~Bengio, ``Density estimation using real
  {NVP},'' \emph{arXiv preprint arXiv:1605.08803}, 2016.

\bibitem{INN_2017understand}
A.~C. Gilbert, Y.~Zhang, K.~Lee, Y.~Zhang, and H.~Lee, ``Towards understanding
  the invertibility of convolutional neural networks,'' in \emph{IJCAI}, 2017,
  pp. 1703--1710.

\bibitem{INN_mintnet}
Y.~Song, C.~Meng, and S.~Ermon, ``Mintnet: Building invertible neural networks
  with masked convolutions,'' in \emph{Adv. Neural Inform. Process. Syst.},
  2019, pp. 11\,004--11\,014.

\bibitem{INN_residualflows}
R.~T. Chen, J.~Behrmann, D.~K. Duvenaud, and J.-H. Jacobsen, ``Residual flows
  for invertible generative modeling,'' in \emph{Adv. Neural Inform. Process.
  Syst.}, 2019, pp. 9916--9926.

\bibitem{est_2018glow}
D.~P. Kingma and P.~Dhariwal, ``{Glow}: Generative flow with invertible 1x1
  convolutions,'' in \emph{Adv. Neural Inform. Process. Syst.}, 2018, pp.
  10\,215--10\,224.

\bibitem{est_2018ffjord}
W.~Grathwohl, R.~T. Chen, J.~Bettencourt, I.~Sutskever, and D.~Duvenaud,
  ``{FFJORD}: Free-form continuous dynamics for scalable reversible generative
  models,'' \emph{arXiv preprint arXiv:1810.01367}, 2018.

\bibitem{inn_2018revnet}
J.-H. Jacobsen, A.~Smeulders, and E.~Oyallon, ``{i-RevNet}: Deep invertible
  networks,'' in \emph{Int. Conf. Learn. Represent.}, 2018.

\bibitem{INN_2019IRB}
J.~Behrmann, W.~Grathwohl, R.~T. Chen, D.~Duvenaud, and J.-H. Jacobsen,
  ``Invertible residual networks,'' in \emph{ICML}, 2019, pp. 573--582.

\bibitem{INN_guided}
L.~Ardizzone, C.~L{\"u}th, J.~Kruse, C.~Rother, and U.~K{\"o}the, ``Guided
  image generation with conditional invertible neural networks,'' \emph{arXiv
  preprint arXiv:1907.02392}, 2019.

\bibitem{rombach2020net2net}
R.~Rombach, P.~Esser, and B.~Ommer, ``{Network-to-Network Translation with
  Conditional Invertible Neural Networks},'' in \emph{Adv. Neural Inform.
  Process. Syst.}, 2020.

\bibitem{Dorkenwald_2021_CVPR}
M.~Dorkenwald, T.~Milbich, A.~Blattmann, R.~Rombach, K.~G. Derpanis, and
  B.~Ommer, ``Stochastic image-to-video synthesis using cinns,'' in \emph{IEEE
  Conf. Comput. Vis. Pattern Recog.}, June 2021, pp. 3742--3753.

\bibitem{sr_2020eccv}
M.~Xiao, S.~Zheng, C.~Liu, Y.~Wang, D.~He, G.~Ke, J.~Bian, Z.~Lin, and T.-Y.
  Liu, ``Invertible image rescaling,'' \emph{Eur. Conf. Comput. Vis.}, 2020.

\bibitem{xie2021enhanced}
Y.~Xie, K.~L. Cheng, and Q.~Chen, ``Enhanced invertible encoding for learned
  image compression,'' in \emph{ACM Int. Conf. Multimedia}, 2021, pp. 162--170.

\bibitem{app_2019residual}
X.~Zhu, Z.~Li, X.-Y. Zhang, C.~Li, Y.~Liu, and Z.~Xue, ``Residual invertible
  spatio-temporal network for video super-resolution,'' in \emph{AAAI},
  vol.~33, 2019, pp. 5981--5988.

\bibitem{liu2021invDN}
Y.~Liu, Z.~Qin, S.~Anwar, P.~Ji, D.~Kim, S.~Caldwell, and T.~Gedeon,
  ``Invertible denoising network: A light solution for real noise removal,'' in
  \emph{IEEE Conf. Comput. Vis. Pattern Recog.}, 2021, pp. 13\,365--13\,374.

\bibitem{zhao2021deep}
L.~Zhao, S.-P. Lu, T.~Chen, Z.~Yang, and A.~Shamir, ``Deep symmetric network
  for underexposed image enhancement with recurrent attentional learning,'' in
  \emph{Int. Conf. Comput. Vis.}, 2021, pp. 12\,075--12\,084.

\bibitem{zhao2021decolor}
R.~Zhao, T.~Liu, J.~Xiao, D.~P. Lun, and K.-M. Lam, ``Invertible image
  decolorization,'' \emph{IEEE Trans. Image Process.}, vol.~30, pp. 6081--6095,
  2021.

\bibitem{shin2017jpeg}
R.~Shin and D.~Song, ``Jpeg-resistant adversarial images,'' in \emph{NIPS 2017
  Workshop on Machine Learning and Computer Security}, vol.~1, 2017.

\bibitem{div2k_2017_CVPR_Workshops}
E.~Agustsson and R.~Timofte, ``Ntire challenge on single image
  super-resolution: Methods and results,'' in \emph{IEEE Conf. Comput. Vis.
  Pattern Recog. Worksh.}, 2017, pp. 114--125.

\bibitem{flick2k_2017_CVPR_Workshops}
B.~Lim, S.~Son, H.~Kim, S.~Nah, and K.~Mu~Lee, ``Enhanced deep residual
  networks for single image super-resolution,'' in \emph{IEEE Conf. Comput.
  Vis. Pattern Recog. Worksh.}, 2017, pp. 136--144.

\bibitem{imagenet_ijcv15}
O.~Russakovsky, J.~Deng, H.~Su, J.~Krause, S.~Satheesh, S.~Ma, Z.~Huang,
  A.~Karpathy, A.~Khosla, M.~Bernstein \emph{et~al.}, ``Imagenet large scale
  visual recognition challenge,'' \emph{Int. J. Comput. Vis.}, vol. 115, no.~3,
  pp. 211--252, 2015.

\end{thebibliography}

\end{document}